\documentclass{sigkddExp}

\begin{document}

\title{Web Mining Research: A Survey}

\numberofauthors{2}
\author{
\alignauthor Raymond Kosala \\
       \affaddr{Department of Computer Science}\\
       \affaddr{Katholieke Universiteit Leuven}\\
       \affaddr{Celestijnenlaan 200A, B-3001 Heverlee, Belgium}\\
       \email{Raymond@cs.kuleuven.ac.be}
\alignauthor Hendrik Blockeel\\
       \affaddr{Department of Computer Science}\\
       \affaddr{Katholieke Universiteit Leuven}\\
       \affaddr{Celestijnenlaan 200A, B-3001 Heverlee, Belgium}\\
       \email{Hendrik.Blockeel@cs.kuleuven.ac.be}
}

\maketitle

\begin{abstract}

With the huge amount of information available online, the World
Wide Web is a fertile area for data mining research. The Web
mining research is at the cross road of research from several
research communities, such as database, information retrieval,
and within AI, especially the sub-areas of machine learning and
natural language processing. However, there is a lot of
confusions when comparing research efforts from different point
of views. In this paper, we survey the research in the area of
Web mining, point out some confusions regarded the usage of the
term Web mining and suggest three Web mining categories. Then we
situate some of the research with respect to these three
categories. We also explore the connection between the Web mining
categories and the related agent paradigm. For the survey, we
focus on representation issues, on the process, on the learning
algorithm, and on the application of the recent works as the
criteria. We conclude the paper with some research issues.

\end{abstract}


\section{Introduction}

The World Wide Web (Web) is a popular and interactive medium to
disseminate information today. The Web is huge, diverse, and dynamic
and thus raises the scalability, multimedia data, and temporal
issues respectively. Due to those situations, we are currently
drowning in information and facing information
overload~\cite{Mae94}. Information users could encounter, among
others, the following problems when interacting with the Web:

a. Finding relevant information: People either browse or use the
search service when they want to find specific information on the
Web. When a user uses search service he or she usually inputs a
simple keyword query and the query response is the list of pages
ranked based on their similarity to the query. However today's
search tools have the following problems \cite{Cha00}. The first
problem is low precision, which is due to the irrelevance of many of
the search results. This results in a difficulty finding the
relevant information. The second problem is low recall, which is due
to the inability to index all the information available on the Web.
This results in a difficulty finding the unindexed information that
is relevant. See \cite{LG99} for some other search engine problems.

b. Creating new knowledge out of the information available on the
Web: Actually this problem could be regarded as a sub-problem of
the problem above. While the problem above is usually a
query-triggered process (retrieval oriented), this problem is a
data-triggered process that presumes that we already have a
collection of Web data and we want to extract potentially useful
knowledge out of it (data mining oriented). Recent research
\cite{CD+98,MB+99,Coh99a} focuses on utilizing the Web as a
knowledge base for decision making.

c. Personalization of the information: This problem is often
associated with the type and presentation of information, since
it is likely that people differ in the contents and presentations
they prefer while interacting with the Web.

On the other hand, the information providers could encounter these
problems, among others, when trying to achieve their goals on the
Web:

d. Learning about consumers or individual users: This is a problem
that specifically deals with the problem c above, which is about
knowing what the customers do and want. Inside this problem,
there are sub-problems such as mass customizing the information
to the intended consumers or even to personalize it to individual
user, problems related to effective Web site design and
management, problems related to marketing, etc.

Web mining techniques could be used to solve the information
overload problems above directly or indirectly. However, we do
not claim that Web mining techniques are the only tools to solve
those problems. Other techniques and works from different research
areas, such as database (DB), information retrieval (IR), natural
language processing (NLP), and the Web document community, could
also be used. By the direct approach we mean that the application
of the Web mining techniques directly addresses the above
problems. For example, a Newsgroup agent that classifies whether
the news is relevant to the user. By the indirect approach we
mean that the Web mining techniques are used as a part of a
bigger application that addresses the above problems. For
example, Web mining techniques could be used to create index
terms for the Web search services.

The Web mining research is a converging research area from several
research communities, such as database, IR, and AI research communities
especially from machine learning and NLP. This paper is an attempt
to put the research done in a more structured way from the machine
learning point of view. However, the methods of the research
that we survey do not necessarily use well-known machine learning
algorithms. Since this is a huge, interdisciplinary, and very
dynamic research area, there are undoubtedly some omissions in
our coverage.

This paper is structured as follows. In section 2 we give an
overview of Web mining, describe some confusions in the usage of
the term Web mining, provide a classification, and relate this
classification to the agent paradigm. In section 3, 4 and 5 we
describe some research that represent the range of the research
in their respective categories. In section 6 we discuss some
related work and finally we conclude in section 7.

\section{Web Mining}

\subsection{Overview}

Web mining is the use of data mining techniques to automatically
discover and extract information from Web documents and services
\cite{Etz96}. This area of research is so huge today partly due to
the interests of various research communities, the tremendous
growth of information sources available on the Web and the recent
interest in e-commerce. This phenomenon partly creates confusion
when we ask what constitutes Web mining and when comparing
research in this area. Similar to Etzioni \cite{Etz96}, we suggest
decomposing Web mining into these subtasks, namely:

\begin{enumerate}

\item Resource finding: the task of retrieving intended Web documents.
\item Information selection and pre-processing: automatically selecting
and pre-processing specific information from retrieved Web resources.
\item Generalization: automatically discovers general patterns at individual
Web sites as well as across multiple sites.
\item Analysis: validation and/or interpretation of the mined patterns.

\end{enumerate}

By resource finding we mean the process of retrieving the data
that is either online or offline from the text sources available
on the Web such as electronic newsletters, electronic newswire,
newsgroups, the text contents of HTML documents obtained by
removing HTML tags, and also the manual selection of Web
resources. We also include text sources that originally were not
accessible from the Web but are accessible now, such as online
texts made for research purposes only, text databases, etc. The
information selection and pre-processing step is any kind of
transformation processes of the original data retrieved in the IR
process. These transformations could be either a kind of
pre-processing that are mentioned above such as removing stop
words, stemming, etc. or a pre-processing aimed at obtaining the
desired representation such as finding phrases in the training
corpus, transforming the representation to relational or first
order logic form, etc. In step 3 above, machine learning or data
mining techniques are typically used for the generalization. We
should also note that humans play an important role in the
information or knowledge discovery process on the Web since the
Web is an interactive medium. This is especially important for
validation and/or interpretation in step 4. Thus, interactive
query-triggered knowledge discovery is as important as the more
automatic data-triggered knowledge discovery. However, we exclude
the knowledge discovery done manually by humans. As we will see
later in section 3, the process 1 - 3 - 4 is also used.

Thus, Web mining refers to the overall process of discovering
potentially useful and previously unknown information or knowledge
from the Web data. It implicitly covers the standard process of
knowledge discovery in databases (KDD) \cite{FSS96b}. We could
simply view Web mining as an extension of KDD that is applied on the
Web data. From the KDD point of view, the information and knowledge
terms are interchangeable \cite{FSS96a}. There is a close
relationship between data mining, machine learning and advanced data
analysis \cite{Mit99}. However, throughout the paper, we discuss the
Web mining research where machine learning techniques are used.
Although mining is an intriguing word to use, it is not a good
metaphor to describe the overall knowledge discovery process
\cite{FSS96a} and what people really do in the field \cite{Hea99}.
Web mining is often associated with IR or IE. However, web mining or
information discovery on the Web is not the same as IR or IE.

\subsubsection{Web Mining and Information Retrieval}

Some have claimed that resource or document discovery (IR) on the
Web is an instance of Web (content) mining and others associate
Web mining with intelligent IR. Actually IR is the automatic
retrieval of all relevant documents while at the same time
retrieving as few of the non-relevant as possible \cite{Van79}.
IR has the primary goals of indexing text and searching for
useful documents in a collection and nowadays research in IR
includes modeling, document classification and categorization,
user interfaces, data visualization, filtering, etc. \cite{YN99}.
The task that can be considered to be an instance of Web mining
is Web document classification or categorization, which could be
used for indexing. Viewed in this respect, Web mining is part of
the (Web) IR process. However, we should note that not all of the
indexing tasks use data mining techniques.

\subsubsection{Web Mining and Information Extraction}

IE has the goal of transforming a collection of documents, usually
with the help of an IR system, into information that is more
readily digested and analyzed \cite{CL96}. IE aims to extract
relevant facts from the documents while IR aims to select relevant
documents \cite{Paz97}. While IE is interested in the structure or
representation of a document, IR views the text in a document
just as a bag of unordered words \cite{Wil97}. Thus, in general IE
works at a finer granularity level than IR does on the documents.
However, the differences between the two become blurred if the
interest of IR is in extraction \cite{Paz99}, and when used in the
context of vague forms of information in which a full text IR
system can provide some IE features \cite{Wil97}.

Building IE systems manually is not feasible and scalable for such
a dynamic and diverse medium such as Web contents \cite{MMK98}.
Due to this nature of the Web, most IE systems focus on specific
Web sites to extract. Others use machine learning or data mining
techniques to learn the extraction patterns or rules for Web
documents semi-automatically or automatically \cite{Kus99}.
Within this view, Web mining is part of the (Web) IE process.
Other views regarding the relationship between (Web) IE and Web
mining also exist. The results of the IE process could be in the
form of a structured database or could be a compression or
summary of the original text or documents. One could view for the
former that IE is a kind of pre-processing stage in the Web
mining process, which is the step after the IR process and before
the data mining techniques are being performed. In a similar
view, IE can also be used to improve the indexing process, which
is part of the IR process. Conversely, one could also argue for
the latter that IE is an instance of text or Web mining since the
summary or the compressed form of a document is a new form of
information that does not exist before. However, we advocate the
view that Web mining is used to improve Web IE (Web mining is
part of IE).

There are basically two types of IE: IE from unstructured texts
and IE from semi-structured data \cite{Mus99}. There are
considerable differences between the IE systems that are used for
unstructured documents with those that are used for
semi-structured or even structured documents. IE tasks from
unstructured natural language texts (classical or traditional IE
tasks) typically use a rather basic to a slightly deeper
linguistic pre-processing before performing data mining.
Classical or traditional IE research, with roots on the NLP
community, has been studied for quite a long time
\cite{CL96,Wil97}. We could say that Advanced Research Projects
Agency (ARPA) helped creating the field (classical IE) because
the evaluations of IE cannot be separated from the ARPA sponsored
Message Understanding Conferences (MUCs) and the TIPSTER IE
project \cite{Wil97,AI99}. MUCs and TIPSTER are competitive
environments that seek to improve IE and IR technologies
\cite{CL96,Car97}. Classical IE usually relies on linguistic
pre-processing such as syntactic analysis, semantic analysis, and
discourse analysis \cite{Sod99,Mus99,Kus99}. Indeed, classical IE
could be called a core language technology \cite{Wil97}.

With the increasing popularity of the Web, there is a need for
structural IE systems that extract information from
semi-structured documents. Structural IE research is different
from the classical one as it usually utilizes the
meta-information (e.g. HTML tags \cite{Sod99}, simple syntactics
\cite{Kus99}, or delimiters \cite{Mus99} that are available
inside the semi-structured data. Structural IE approaches that do
not use linguistic constraints are termed wrapper induction
\cite{Mus99}. Some of the structural IE systems are built manually
by knowledge engineering approach, for examples see
\cite{CM+94,AM97,HG+97}. However, more and more structural IE
systems for the Web are built (semi-) automatically using machine
learning techniques or other algorithms as building the systems
manually is no longer appropriate \cite{Kus99}. Some examples are
\cite{KWD97,Fre98,HD98,MMK98,GM99,Sod99}. These systems are
usually built by using machine learning or data mining
techniques, which learn extraction rules from the annotated
corpora. For more explanations and the categories of IE we point
interested readers to the following survey papers. For classical
IE and the issues of IE for unstructured texts we refer to
\cite{CL96,Car97,AI99,Sod99} and for structural IE we refer to
\cite{Sod99, Mus99}.

\subsubsection{Web Mining and Machine Learning Applied on the Web}

Web mining is not the same as learning from the Web or machine
learning techniques applied on the Web. On the one hand, there
are some applications of machine learning applied on the Web that
are not instances of Web mining. An example of this is a machine
learning technique that is used to spider the Web efficiently for
a specific topic \cite{RM99a,MN+99} that emphasize on planning the
best path that is going to be traversed next. On the other hand,
there are some other methods used for Web mining besides machine
learning methods. Some examples are some proprietary algorithms
that are used for mining the hubs and authorities \cite{CD+99},
DataGuides \cite{GW97, GW99} and Web schema discovery \cite{WL97,
NAM98}. However, there is a close relationship between the two
research areas. Machine learning techniques support and help Web
mining as they could be applied to the processes in Web mining.
For example recent research \cite{Mla99} shows that applying
machine learning techniques could improve the text classification
process compared to the traditional IR techniques. In short, Web
mining intersects with the application of machine learning on the
Web.

\subsection{Web Mining Categories}

In this section we give the overview of each category. More
detailed explanations are given in the respective sections.
Similar to Madria, et al. \cite{MB+99} and Borges and Levene
\cite{BL99}, we categorize Web mining into three areas of interest
based on which part of the Web to mine: Web content mining, Web
structure mining, and Web usage mining. Web content mining
describes the discovery of useful information from the Web
contents/data/documents. However, what consist of the Web
contents could encompass a very broad range of data. Previously
the Internet consists of different types of services and data
sources such as Gopher, FTP and Usenet. Now most of those data are
either ported to or accessible from the Web. It is mentioned in
\cite{HC+99} that in the last several years the growth in the
amount of government information has been tremendous. We also
know the existence of Digital Libraries that are also accessible
from the Web. We also see that many companies are transforming
their businesses and services electronically. As a consequence
many of the company databases that previously resided in the
legacy systems are being ported to or made accessible from the
Web. Thus the employees, partners, or even customers could access
some of the company database directly from Web based interfaces.
Another consequence of this transformation is the existence of Web
applications so that the users could access the applications
through Web interfaces. Many applications and systems are being
migrated to the Web and many types of applications are emerging
in the Web environment. Of course some of the Web content data
are hidden data, which cannot be indexed. These data are either
generated dynamically as a result of queries and reside in the
DBMSs or are private. In short, the Web already contains many
kinds and types of data.

Basically, the Web content consists of several types of data such as
textual, image, audio, video, metadata as well as hyperlinks. Recent
research on mining multi types of data is termed multimedia data
mining \cite{ZH+98}. Thus we could consider multimedia data mining
as an instance of Web content mining. However this line of research
still receives less attention than the research on the text or
hypertext contents \cite{ZH+98,Mit99}. The Web content data consist
of unstructured data such as free texts, semi-structured data such
as HTML documents, and a more structured data such as data in the
tables or database generated HTML pages. However, much of the Web
content data is unstructured text data
\cite{Etz96,CC+98,AH+99,Cha00}. The research around applying data
mining techniques to unstructured text is termed knowledge discovery
in texts (KDT) \cite{FD95}, or text data mining \cite{Hea99}, or
text mining \cite{Tan99}. Hence we could consider text mining as an
instance of Web content mining. We discuss text mining further in
the next section. We could differentiate the research done in Web
content mining from two different points of view: IR and DB
\cite{CMS97} views. The goal of Web content mining from the IR view
is mainly to assist or to improve the information finding or
filtering the information to the users usually based on either
inferred or solicited user profiles, while the goal of Web content
mining from the DB view mainly tries to model the data on the Web
and to integrate them so that more sophisticated queries other than
the keywords based search could be performed. These viewpoints are
further discussed in the next section.

Web structure mining \cite{CD+99} tries to discover the model
underlying the link structures of the Web. The model is based on the
topology of the hyperlinks with or without the description of the
links. This model can be used to categorize Web pages and is useful
to generate information such as the similarity and relationship
between different Web sites. Web structure mining could be used to
discover authority sites for the subjects (authorities) and overview
sites for the subjects that point to many authorities (hubs).

Web usage mining \cite{CMS97} tries to make sense of the data
generated by the Web surfer's sessions or behaviors. While the
Web content and structure mining utilize the real or primary data
on the Web, Web usage mining mines the secondary data derived
from the interactions of the users while interacting with the
Web. The Web usage data includes the data from Web server access
logs, proxy server logs, browser logs, user profiles,
registration data, user sessions or transactions, cookies, user
queries, bookmark data, mouse clicks and scrolls, and any other
data as the results of interactions. Table 1 gives an overview of
the above Web mining
categories (the explanations are given in the subsequent sections).

However we should emphasize that the distinctions between the
above categories are not clear-cut. Web content mining might
utilize text and links and even the profiles that are either
inferred or inputted by the users. User profiles are mostly used
for the user modeling applications or personal assistants. The
same is true for Web structure mining that could use the
information about the links in addition to the link structures.
Moreover we could infer the traversed links from the documents
that were requested during the user session from the logs
generated by the server. We could also characterize the
categories above from the point of view of the \textit{scope} of
most of the work done in the respective areas: local scope spans
an individual Web site while global scope spans the entire Web.
The scope of the Web content mining from the IR view and Web
structure mining is global while the scope of the Web content
mining from the DB view and Web usage mining is local. However
this characterization is not clear-cut either.

\begin{table*}
\centering
\scriptsize
\caption{Web mining categories}
\begin{tabular}{|p{0.74in}|p{1.4in}|p{1.4in}|p{1.3in}|p{1.4in}|} \hline

& \multicolumn{4}{c|}{Web Mining} \\ \cline{2-5}
& \multicolumn{2}{c|}{Web Content Mining} &
  \multicolumn{1}{c|}{Web Structure Mining} &
  \multicolumn{1}{c|}{Web Usage Mining} \\ \cline{2-3}
& \multicolumn{1}{c|}{IR View} & \multicolumn{1}{c|}{DB View} & & \\ \hline

View of Data & - Unstructured & - Semi structured &
  - Links structure & - Interactivity \\
& - Semi structured & - Web site as DB & & \\ \hline

Main Data & - Text documents & - Hypertext documents &
- Links structure & - Server logs \\
& - Hypertext documents & & & - Browser logs \\ \hline

Representation & - Bag of words, n-grams
& - Edge-labeled graph (OEM) & - Graph & - Relational table \\
& - Terms, phrases & - Relational & & - Graph \\
& - Concepts or ontology & & & \\
& - Relational & & & \\ \hline

Method & - TFIDF and variants & - Proprietary algorithms &
- Proprietary algorithms & - Machine Learning \\
& - Machine learning & - ILP & & - Statistical \\
& - Statistical (including NLP) & - (Modified) association rules & &
- (Modified) association rules \\ \hline

Application Categories & - Categorization & - Finding frequent sub-structures &
- Categorization & - Site construction, adaptation, and management \\
& - Clustering & - Web site schema discovery & - Clustering & - Marketing \\
& - Finding extraction rules & & & - User modeling \\
& - Finding patterns in text & & & \\
& - User modeling & & & \\ \hline

\end{tabular}
\end{table*}

In practice, the three Web mining tasks above could be used in
isolation or combined in an application, especially in Web content
and structure mining since the Web documents might also contain
links. For example, Chakrabarti et al. \cite{CD+99} uses as Web
content the terms in a document's link neighborhood and as Web
structure the links from its neighbors, to classify Web pages.
Joachims et al. \cite{JFM97} use Web content and usage to build a
software tour agent for assisting users browsing a Web site.

\subsection{Web Mining and the Agent Paradigm}

Web mining is often viewed from or implemented within an agent
paradigm. Thus, Web mining has a close relationship with software
agents or intelligent agents. Indeed some of these agents perform
data mining tasks to achieve their goals. According to Green, et
al. \cite{GH+97} there are three sub-categories of software
agents: user interface agents, distributed agents, and mobile
agents. The sub-categories of software agents that are relevant
for data mining tasks are user interface agents and distributed
agents. User interface agents try to maximize the productivity of
current users interaction with the system by adapting behavior.
The issue of personalization abounds here. User interface agents
that can be classified into the Web mining agent category are
information retrieval agents, information filtering agents, and
personal assistant agents. Distributed agents technology is
concerned with problem solving by a group of agents and relevant
agents in this category are distributed agents for knowledge
discovery or data mining (for example see \cite{KHS97}).

There are two frequently used approaches for developing
intelligent agents that help users find and retrieve relevant
information from the Web \cite{BS97}, namely content-based and
collaborative approaches. In the content-based approach, the
system searches for items that match based on an analysis of the
content using the user preferences. In the collaborative approach,
the system tries to find users with similar interests to give
recommendations to. The system does this by analyzing the user
profiles and sessions or transactions. It assumes that if some
users rate an item high, then the other users with similar
interests would rate this item high also. So this approach mainly
uses the usage data (user ratings). Viewed in this light we could
categorize the content-based methods as Web content mining and
categorize the collaborative approaches as Web usage mining.
However, collaborative approaches
might also be used or combined with the Web content.

A similar view related to the Web mining categories above also
exists in the software agent community. Delgado \cite{Del00}
classifies the user interface agents by the underlying
information filtering technology into content-based filters,
reputation based filters, collaborative or social-based filters,
event-based filters, and hybrid filters. In event based
filtering, the system tracks and follows the events that are
inferred from the surfing habits of people in the Web. Some
examples of those events are saving a URL into a bookmark folder,
mouse clicks and scrolls, link traverse behavior, etc. We could
make an association between these agent-based categories with the
Web mining categories above. Table 2 shows the association.

\begin{table}
\scriptsize \centering \caption{The association between the
categories of Web mining and the agent paradigm}
\begin{tabular}{p{1.4in}p{0.1in}p{1.5in}}
\space & \space & \space \\
Content-based filters & $\leftrightarrow$ & Content mining\\
Reputation-based filters & $\leftrightarrow$ & Structure (and content) mining\\
Collaborative or social-based filters & $\leftrightarrow$ & Usage mining\\
Event-based filters & $\leftrightarrow$ & Usage mining\\
Hybrid filters & $\leftrightarrow$ & Combination of the categories\\
\end{tabular}
\end{table}

\section{Web Content Mining}

In this section we list some of the research in the respective
categories in separate tables. We should note that the lists are
by no means complete. The explanations on the methods surveyed
are beyond the scope of this paper. Interested readers can consult
the book by Mitchell \cite{Mit97} and the respective papers for
the explanation of the methods. We just intend to give a taste on
the variety of some representations, processes, methods, and
applications that have been used.

\subsection{Information Retrieval View}

\subsubsection{Information Retrieval View for Unstructured Documents}

Table 3 summarizes some of the research done for unstructured
documents. What we mean by the unstructured documents is free
texts such as news stories. Most of the research in table 3 uses
bag of words to represent unstructured documents. The bag of
words or vector representation \cite{SM83} takes single words
found in the training corpus as features. This representation
ignores the sequence in which the words occur and is based on the
statistic about single words in isolation. The features could be
Boolean (a word either occurs or does not occur in a document),
or frequency based (frequency of the word in a document).
Variations of the feature selection include removing the case,
punctuation, infrequent words, and stop words. The features could
be reduced further by applying some other feature selection
techniques, such as information gain, mutual information, cross
entropy, or odds ratio (see \cite{MG99} for the details). Other
preprocessing includes Latent Semantic Indexing (LSI) \cite{DD+90}
that seeks to transform the original document vectors to a lower
dimensional space by analyzing the correlational structure of
terms in the document collection such that similar documents that
do not share terms are placed in the same topic, and stemming
which reduces words to their morphological roots. For example the
words ``informing'', ``information'', ``informer'', and
``informed'' would be stemmed to their common root ``inform'' and
only the latter word is used as the feature instead of the former
four. While those pre-processing variations are useful for
reducing feature set size, the generality of their effectiveness
over different domains for text categorization tasks are doubted
\cite{Ril95}.

\begin{table*}
\scriptsize \caption{An IR view on Web content mining for
unstructured documents}
\begin{tabular}{|p{1.2in}|p{1.1in}|p{0.6in}|p{1.5in}|p{1.5in}|} \hline

Author & Document Representation & Process & Method & Application
\\ \hline

Ahonen, et al. \cite{AH+98} & Bag of words and word positions & 1
- 2
- 3 - 4 & Episode rules & - Finding keywords and keyphrases \\
& & & & - Discovering grammatical rules and collocations \\ \hline

Billsus and Pazzani \cite{BP99} & Bag of words & 1 - 2 - 3 - 4 &
- TFIDF & Text classification \\
& & & - Na\"{\i}ve Bayes & \\ \hline

Cohen \cite{Coh95} & Relational & 1 - 2 - 3 - 4 &
- Propositional rule based system & Text classification \\
& & & Inductive Logic Programming & \\ \hline

Dumais, et al. \cite{DP+98} & - Bag of words & 1 - 2 - 3 - 4 &
- TFIDF & Text categorization \\
& - Phrases & & - Decision trees & \\
& & & - Naive Bayes & \\
& & & - Bayes nets & \\
& & & - Support Vector Machines & \\ \hline

Feldman and Dagan \cite{FD95} & Concept categories & 1 - 2 - 3 - 4
& Relative entropy & Finding patterns between concept
distributions in textual data
\\ \hline

Feldman, et al. \cite{FF+98} & Terms & 1 - 2 - 3 - 4 & Association
rules & Finding patterns across terms in textual data \\ \hline

Frank, et al. \cite{FP+99} & Phrases and their positions & 1 - 2 -
3 -
4 & Naive Bayes & Extracting keyphrases from text documents \\
\hline

Freitag and McCallum \cite{FM99} & Bag of words & 1 - 3 - 4 &
Hidden Markov Models & Learning extraction models \\ \hline

Hofmann \cite{Hof99} & Bag of words & 1 - 2 - 3 - 4 &
\textnormal{Unsupervised} statistical method & Hierarchical
clustering
\\ \hline

Honkela, et al. \cite{HK97} & Bag of words with n-grams & 1 - 2 -
3 - 4 & Self-Organizing Maps & Text and document clustering \\
\hline

Junker, et al. \cite{JSR99} & Relational & 1 - 2 - 3 - 4 &
Inductive Logic Programming & - Text categorization \\
& & & & - Learning extraction rules \\ \hline

Kargupta, et al. \cite{KHS97} & Bag of words with n-grams & 1 - 2
- 3 - 4 & - Unsupervised hierarchical clustering &
Text classification and hierarchical clustering \\
& & & - Decision trees & \\
& & & - Statistical analysis & \\ \hline

Nahm and Mooney \cite{NM00} & Bag of words & 1 - 2 - 3 - 4 &
Decision trees & Predicting (words) relationship \\ \hline

Nigam, et al. \cite{NLM99} & Bag of words & 1 - 3 - 4 & Maximum
entropy & Text classification \\ \hline

Scott and Matwin \cite{SM99} & - Bag of words & 1 - 2 - 3 - 4 &
Rule based system & Text classification \\
& - Phrases & & & \\
& - Hypernyms and synonyms & & & \\ \hline

Soderland \cite{Sod99} & Sentences, and clauses & 1 - 2 - 3 - 4 &
Rule learning & Learning extraction rules \\ \hline

Weiss, et al. \cite{WA+99} & Bag of words & 1 - 2 - 3 - 4 &
Boosted decision trees & Text categorization \\ \hline

Wiener, et al. \cite{WPW95} & Bag of words & 1 - 2 - 3 - 4 &
- Neural Networks & Text categorization \\
& & & - Logistic Regression & \\ \hline

Witten, et al. \cite{WB+99} & Named entity & 1 - 2 - 3 - 4 & Text
compression & Named entity classifier \\ \hline

Yang, et al. \cite{YC+99} & Bag of words and phrases & 1 - 2 - 3 -
4 &
- Clustering algorithms & Event detection and tracking \\
& & & - k-Nearest Neighbor & \\
& & & - Decision tree & \\ \hline

\end{tabular}
\end{table*}

Other feature representations are also possible such as using
information about word positions in the document
\cite{Coh95,AH+98,FP+99}, using n-grams representation (word
sequences of length up to n) \cite{HK97,KHS97} (for example ``the
morphological roots'' is a tri-gram), using phrases
\cite{DP+98,FP+99,SM99,YC+99} such as ``the quick brown fox that
run away'', using document concept categories \cite{FD95}, using
terms \cite{FF+98} such as ``annual interest rate'' or ``Wall
Street'', using hypernyms (linguistic term for the ``is a''
relationship - a dog is an animal, thus ``animal'' is a hypernym
of ``dog'') \cite{SM99}, or using named entities \cite{WB+99} such
as people's names, dates, email addresses, locations,
organizations, or URLs. The relational representation
(\cite{Coh95,JSR99} in table 3) that we mean here is actually
first order logic, a language that is more expressive than
propositional logic (for instance see \cite{Mit97}). For example
in the bag of words representation features are the frequencies of
specific words; using a relational representation one might use
relationships between different words and their positions, e.g.
``word X is to the left of word Y in the same sentence''.
Although different types of representations have been used, there
is currently no study that shows clear advantages of some
representations over several domains for text categorization
tasks \cite{Mla99}. Indeed, Scott and Matwin \cite{SM99} compare
different representations (bag of words, phrase based, and
hypernym) but found no significant differences in the performance
of different representations.

As we can see from table 3, the commonly used process is 1 - 2 - 3 -
4, while some others do not use any or only use a minimal
pre-processing step (process 1 - 3 - 4). The name and explanation of
the four steps are described in section 2.1 above. The use of text
compression techniques \cite{WB+99} is rather new for the text
classification task. The applications range from text classification
or categorization, event detection and tracking, finding extraction
patterns or rules, to finding some interesting patterns in the text
documents. Event detection and tracking problems are sub-topics of a
broader initiative called topic detection and tracking (TDT), which
is a new line of research related to research in information
retrieval and filtering \cite{APL98}. TDT is an initiative to
investigate the state of the art in finding and following new events
in a stream of news stories broadcast \cite{AC+98}.

Recently the usage of the term text mining has been a subject to
controversy. There are at least two controversies that we are
aware of: one is regarding the usage of the term ``mining''
itself \cite{Hea99} and the other one is regarding the meaning of
the word ``knowledge'' in knowledge discovery from text (KDT)
\cite{Kod99}. As far as we know, the term text mining or KDT was
first proposed by Feldman and Dagan in \cite{FD95}. They suggest
to structure the text documents by means of information
extraction, text categorization, or applying NLP techniques as
pre-processing step before performing any kind of KDTs. The
reason is mining on the unprepared documents does not provide
effectively exploitable results \cite{RB98,FF+98}. Currently the
term text mining has been used to describe different applications
such as text categorization \cite{HW99,Tan99,WA+99}, text
clustering \cite{Tan99,RM99b}, IE \cite{AH+98}, empirical
computational linguistic tasks \cite{Hea99}, exploratory data
analysis \cite{Hea99}, finding patterns in text databases
\cite{FD95,FF+98}, finding sequential patterns in texts
\cite{LAS97,AH+98,AH+99}, and association discovery
\cite{Tan99,NM00}. So although some of the papers surveyed mention
their application as text mining, we use less controversial names
for the applications.

\begin{table*}
\scriptsize \centering \caption{An IR view on Web content mining
for semi-structured documents}
\begin{tabular}{|p{1.2in}|p{1.1in}|p{0.6in}|p{1.5in}|p{1.5in}|} \hline

Author & Document Representation & Process & Method & Application
\\ \hline

Craven, et al. \cite{CD+98} & Relational and ontology & 1 - 2 - 3
- 4 &
- Modified Naive Bayes & - Hypertext classification \\
& & & - Inductive Logic Programming & - Learning Web page relation \\
& & & & - Learning extraction rules \\ \hline

Crimmins, et al. \cite{CS+99} & Phrase, URLs, and meta information
& 1 - 2 - 3 - 4 & Unsupervised and supervised classification
algorithms & - Hierarchical and graphical classification \\
& & & & - Clustering \\ \hline

F\"{u}rnkranz \cite{Fur99} & Bag of words and hyperlinks
information & 1 - 2 - 3 - 4 & Rule learning & Hypertext
classification \\ \hline

Joachims, et al. \cite{JFM97} & Bag of words and hyperlinks
information & 1 - 2 - 3 - 4 & - TFIDF & Hypertext prediction \\
& & & - Reinforcement learning & \\ \hline

Muslea, et al. \cite{MMK98} & Bag of words, tags, and word
positions &
1 - 2 - 3 - 4 & Rule learning & Learning extraction rules \\
\hline

Shavlik and Eliassi-Rad \cite{SR98} & Localized bag of words, and
relational. & 1 - 2 - 3 - 4 & Neural networks with reinforcement
learning & Hypertext (homepage) classification \\ \hline

Singh, et al. \cite{SC+98} & Concepts and Named entity & 1 - 2 - 3
- 4 & - Modified association rule & Finding patterns in
semi-structured texts \\
& & & - Classification algorithm & \\ \hline

Soderland \cite{Sod99} & Sentences, phrases, and named entity & 1
- 2 - 3 - 4 & Rule learning & Learning extraction rules \\ \hline

\end{tabular}
\end{table*}

\subsubsection{Information Retrieval View for Semi-Structured Documents}

We can see from table 4 that the process used in the works surveyed
above is 1 - 2 - 3 - 4. We can also see that the works surveyed
in table 4 use richer representations compared to the works surveyed
in table 3. This is due to the additional structural (HTML and
hyperlink) information in the hypertext documents. Actually all
of the works surveyed utilize the HTML structures inside the
documents and some utilize the hyperlink structure between the
documents for document representation. The methods that are used
are common data mining methods. The applications ranged from
hypertext classification or categorization and clustering, learning
relations between Web documents, learning extraction patterns
or rules, and finding patterns in semi-structured data.

\subsection{Database View}

As mentioned in \cite{FLM98}, the database techniques on the Web
are related to the problems of managing and querying the
information on the Web. \cite{FLM98} mentions that there are three
classes of tasks related to those problems: modeling and querying
the Web, information extraction and integration, and Web site
construction and restructuring. Although the first two tasks are
related to Web content mining applications, not all the works
there are inside the scope of Web content mining. This is due to
the absence of the machine learning or data mining techniques in
the process. Basically the DB view tries to infer the structure
of the Web site or to transform a Web site to become a database
so that better information management and querying on the Web
become possible. As mentioned previously, the DB view of Web
content mining mainly tries to model the data on the Web and to
integrate them so that more sophisticated queries other than the
keywords based search could be performed. These could be achieved
by finding the schema of Web documents, building a Web warehouse
or a Web knowledge base or a virtual database. The research done
in this area mainly deals with semi-structured data.
Semi-structured data from database view often refers to data that
has some structure but no rigid schema \cite{Abi97,Bun97}.

\begin{table*}
\scriptsize \centering \caption{Web content mining from a database
view}
\begin{tabular}{|p{1.2in}|p{1.1in}|p{0.6in}|p{1.5in}|p{1.5in}|} \hline

Author & Document Representation & Process & Method & Application
\\ \hline

Goldman and Widom \cite{GW99} & OEM & 1 - 2 - 3 - 4 & Proprietary
algorithms & Finding DataGuide in semi-structured data \\ \hline

Grumbach and Mecca \cite{GM99} & Strings and relational & 1 - 2 -
3 - 4 & Proprietary algorithms & Finding schema in
semi-structured data
\\ \hline

Nestorov, et al. \cite{NAM97} & OEM & 1 - 2 - 3 - 4 & Proprietary
algorithms & Finding type hierarchy in semi-structured data
\\ \hline

Toivonen \cite{Toi99} & OEM & 1 - 2 - 3 - 4 & Upgraded association
rules & Finding useful sub-structure in semi-structured data
\\ \hline

Wang and Liu \cite{WL99} & OEM & 1 - 2 - 3 - 4 & Modified
association rules & Finding frequent sub-structures in
semi-structured data
\\ \hline

Zaiane and Han \cite{ZH98} & Relational & 1 - 2 - 3 - 4 &
Attribute-oriented induction & Multilevel databases
\\ \hline

\end{tabular}
\end{table*}

From table 5, we can see that the DB view uses representations
that differ from the IR view that we see in table 3 and table 4.
The DB view mainly uses Object Exchange Model (OEM) \cite{AQ+97}
that represents semi-structured data by a labeled graph. The data
in the OEM is viewed as a graph, with objects as the vertices and
labels on the edges. Each object is identified by an object
identifier (oid) and a value that is either atomic, such as
integer, string, gif, html, etc. or complex in the form of a set
of object references, denoted as a set of (label, oid) pairs. All
of the processes that are surveyed above are 1 - 2 - 3 - 4.
However, the process used here typically starts from manually
selected Web sites for doing Web content mining instead of
searching the whole Internet for the specific resources. This is
partly due to the applications of the DB view that are quite
different from those of the IR view (which mostly are
classification tasks). The process 1 and 2 is typically done by
site-specific wrappers or parsers for hypertext documents.

Most of the applications that are surveyed above are the task of
schema extraction or discovery \cite{WL99,Toi99} or building
DataGuides \cite{GW97,NAM97,GW99}. Roughly speaking, a schema or
DataGuide is a kind of structural summary of semi-structured data.
For practical applications and computational reason, this summary is
often approximated \cite{Abi97,GW99}. Some applications do not deal
with the task of finding the global schema but deal with the task of
finding frequent substructures (sub-schema) in semi-structured data.
Another application deals with the task of creating multi-layered
database (MLDB) \cite{ZH98} in which each layer is obtained by
generalizations on lower layers and use a special purpose query
language for Web mining to extract some knowledge from the MLDB of
Web documents. This is an example of the query perspective of data
mining. There has been some work on query languages for
semi-structured data \cite{AQ+97,BD+96} and for the Web
\cite{AM99,LSS96,MMM96,FF+97}. However, we only see the works by
Za\"{\i}ane, et al. \cite{ZH98} and Singh, et al. \cite{SC+98} that
are inside the scope of Web content mining.

Due to the different representations used in the DB view, most of
the methods used for data mining are also different except the
ILP methods that could operate on relational or graphical data.
These differences are partly due to the inappropriateness of many
existing data mining techniques, which operate on flat data, to
operate on relational or graphical data. \cite{GM99,NAM97,ZH98}
use proprietary algorithms for schema discovery and for the
construction of MLDB, \cite{WL99} uses a modified version of
association rules, and \cite{Toi99} uses an upgraded first order
logic version of association rules \cite{DD97}.

\subsection{About Mining Multimedia Data}

We should note that we have not actually discussed the issue of
mining multimedia data on the Web. Although multimedia data has
been the major focus for many researchers \cite{KB96,Sub98} and
many techniques for multimedia IR and extraction have been
proposed (for example see \cite{Hau99}), multimedia data mining is
still in its infancy \cite{ZH+98}. Uthurusamy \cite{Uth96},
Shapiro et al. \cite{SB+96}, and Mitchell \cite{Mit99} assert that
working towards a unifying framework for representation, problem
solving, and learning from multimedia data is indeed a challenge.
Fayyad et al. \cite{FDW96} describes mining the images of sky
objects taken from satellite. Smyth, et al. \cite{SF+96} describes
mining images to identify small volcanoes on Venus. More recent
works are \cite{ZH+98} in the application of Web data warehousing
and \cite{HC+99} in the application of a medical IR system for
mining the multimedia data on the Web. For a definition and a
short survey on multimedia data mining, we refer to \cite{ZH+98}.

\section{Web Structure Mining}

If in the database view of Web content mining we are interested
in the structure within Web documents (intra-document structure),
in Web structure mining we are interested in the structure of the
hyperlinks within the Web itself (inter-document structure). This
line of research is inspired by the study of social networks and
citation analysis \cite{KSS97,Cha00}. With social network analysis
we could discover specific types of pages (such as hubs,
authorities, etc.) based on the incoming and outgoing links. Web
structure mining utilizes the hyperlinks structure of the Web to
apply social network analysis to model the underlying links
structure of the Web itself. Research done by Kautz et al.
\cite{KSS97} utilizes the network analysis of people to model the
network of AI researchers. They use the name entity data found in
close proximity in any public Web pages such as the hyperlinks
from home pages, co-authorship and citation of papers, exchange of
information between individuals found in net-news archives, and
organization charts. In our framework, their research could be
classified as a combination of Web structure and content mining.

Some algorithms have been proposed to model the Web topology such as
HITS \cite{Kle98}, PageRank \cite{BP98} and improvements of HITS by
adding content information to the links structure \cite{CD+99} and
by using outlier filtering \cite{BH98}. These models are mainly
applied as a method to calculate the quality rank or relevancy of
each Web page. Some examples are the Clever system \cite{CD+99} and
Google \cite{BP98}. Some other applications of the models include
Web pages categorization \cite{CDI98} and discovering micro
communities on the Web \cite{KR+98}.

More applications of Web structure mining in the context of Web
warehouse are discussed by Madria, et al. \cite{MB+99}. These
include measuring the completeness of the Web sites by measuring the
frequency of local links that reside in the same server, measuring
the replication of Web documents across the Web warehouse that helps
in identifying the mirrored sites for example, and discovering the
nature of the hierarchy of hyperlinks in the Web sites of a
particular domain to study how the flow of information affects the
design of the Web sites.

\section{Web Usage Mining}

Web usage mining focuses on techniques that could predict user
behavior while the user interacts with the Web. As mentioned
before, the mined data in this category are the secondary data on
the Web as the result of interactions. These data could range
very widely but generally we could classify them into the usage
data that reside in the Web clients, proxy servers and servers
\cite{SC+00}. The Web usage mining process could be classified
into two commonly used approaches \cite{BL99}. The first approach
maps the usage data of the Web server into relational tables
before an adapted data mining technique is performed. The second
approach uses the log data directly by utilizing special
pre-processing techniques. As is true for typical data mining
applications, the issues of data quality and pre-processing are
also very important here. The typical problem is distinguishing
among unique users, server sessions, episodes, etc. in the
presence of caching and proxy servers \cite{MS00,SC+00}. For the
details and comparison of some pre-processing methods for Web
usage data we refer to \cite{CMS99}.

In general, typical data mining methods (see for example in \cite{
SC+00}) could be used to mine the usage data after the data have
been pre-processed to the desired form. However, modifications of
the typical data mining methods are also used such as composite
association rules \cite{BL98},  an extension of a traditional
sequence discovery algorithm (MIDAS \cite{BB+99}), and hypertext
probabilistic grammars \cite{BL99}. The Web usage data could also
be represented with graphs \cite{BB+99,PP+00}. Often the Web
usage mining uses some background or domain knowledge such as
navigation templates, Web content, site topology, concept
hierarchies, and syntactic constraints \cite{BB+99,Spi99}.

The applications of Web usage mining could be classified into two
main categories: learning a user profile or user modeling in
adaptive interfaces (personalized) (for examples see \cite{Lan99})
and learning user navigation patterns (impersonalized) (for examples
see \cite{Spi99}). Web users would be interested in, among others,
techniques that could learn their information needs and preferences,
which is user modeling possibly combined with Web content mining. On
the other hand, information providers would be interested in, among
others, techniques that could improve the effectiveness of the
information on their Web sites by adapting the Web site design or by
biasing the user's behavior towards satisfying the goals of the
site. In other words, they are interested in learning user
navigation patterns. Then the learned knowledge could be used for
applications such as personalization (at a Web site level), system
improvement, site modification, business intelligence, and usage
characterization (see \cite{SC+00} for the detail). It is not in our
intention to give a complete survey of Web usage mining research
here. Interested readers could consult the overview papers by
Srivastava, et al. \cite{SC+00}, Spiliopoulou \cite{Spi99}, and
Masand and Spiliopoulou \cite{MS00}, and Robert Cooley's Ph.D. thesis
\cite{Coo00} for mining user patterns and the overview paper by
Langley \cite{Lan99} for mining user profiles.

\section{Related Works}

As far as we know, it was Etzioni \cite{Etz96} who first coined
the term Web mining. Etzioni starts by making a hypothesis that
the information on the Web is sufficiently structured and outlines
the subtasks of Web mining. His paper describes the Web mining
processes. There have been some works around the survey of data
mining on the Web. The first paper that we know that noticed the
confusion in the Web mining research is \cite{CMS97}. It gives a
Web mining taxonomy but restricted to Web content and Web usage
mining, and gives a survey on Web usage mining. It divides the
Web content mining into the agent based approach and the database
approach. We use a similar division but divide it into the IR
approach instead of the agent approach. Later, in \cite{SC+00}
they classify Web mining into three categories that are similar
to our categories. Compared to their paper, our paper points out
three confusions on the usage of the term Web mining, identifies
additional user-centered Web mining processes, and provides new
perspectives for the Web mining categories. We use the Web mining
categories suggested in \cite{MB+99} and \cite{BL99}. \cite{BL99}
proposes a new model for mining Web log data, while \cite{MB+99}
discusses the research issues of Web mining in the context of Web
warehouse project.

Carbonell et al. \cite{CC+98} give an overview of the workshop on
learning from text and the Web that is related to Web content (from
the IR view) and usage mining. They also give an outline of the
research directions in that area. Mladenic \cite{Mla99} surveys the
research on text learning and related intelligent agents. She
compares two frequently used approaches for developing intelligent
agents, namely collaborative and content based. In our categories,
these would be Web content (from the IR view) and usage mining. She
also surveys research on machine learning applied to text data,
which is broader than but similar to our discussion in section 3.1.1
about the IR view of Web content mining from unstructured documents.
Carbonell et al. \cite{CYC00} review the emerging research
collaborations between the IR and machine learning communities in a
special issue of the Machine Learning journal. They also indicate
some fertile research areas for both communities. Garofalakis et al.
\cite{GR+99} review some data mining techniques and the algorithms
for Web mining that specifically take into account the hyperlink
information. Chakrabarti \cite{Cha00} provides a survey of data
mining for hypertext. His paper mainly surveys the statistical
techniques for Web content across the continuum of supervised,
semi-supervised and unsupervised learning, and social network
analysis techniques for Web structure mining. Levy and Weld
\cite{LW00} wrote a survey in the special issue of Artificial
Intelligence on intelligent Internet systems that we think describes
a broader domain than Web mining. Vaithyanathan \cite{Vai99} gives
an overview of the papers in the special issue of Artificial
Intelligence Review on data mining on the Internet. He mentions
similar categories of Web mining as ours, except the database view
of Web content mining. Some other related work that we found
recently in special issues of some magazines are the following. Yang
and Pedersen edited a special issue on intelligent information
retrieval \cite{YP99}. Filman and Pant edited a special issue on
searching the Internet \cite{FP99}.

\section{Conclusions}

In this paper we survey the research in the area of Web mining.
We point out some confusions regarded the usage of the term Web
mining. We also suggest three Web mining categories and then
situate some of the research with respect to these categories. We
also explore the connection between Web mining categories and the
related agent paradigm. For the survey, we focus on
representation issues, on the process, and on the learning
algorithm, and the application of the recent works as the
criteria. The Web presents new challenges to the traditional data
mining algorithms that work on flat data. We have seen that some
of the traditional data mining algorithms have been extended or
new algorithms have been used to work on the Web data.

An interesting direction of Web content mining is the recent
interest in information integration \cite{CM+94,FK+00}, which
could be in the form of a Web knowledge base \cite{CC+98,Coh99a}
or Web warehouse \cite{MB+99}, or in the form of a mediator (see
\cite{FK+00} for examples). At least this is the area where
database and other research communities such as IR, AI, and
machine learning met recently. Information integration was mainly
concerned with integrating various databases but has changed its
focus with the increasing popularity of the Web \cite{FK+00}. The
same is also true for the research in IE, which could be thought
as a mediator or wrapper in the information integration area.
Information integration also raises some other research questions
such as scaling up the number of Web sites that could be
integrated, wrapper maintenance, building and maintaining a
global schema, etc. \cite{Coh99b} (see also \cite{Kus99} for other
issues).

Topic detection and tracking (TDT) is also a promising research
area for IR and machine learning communities that raises, among
others, temporal issue in the data. It would be interesting if the
learning algorithm could model this aspect accurately. Some other
promising research issues in the area of Web content mining are
discussed in \cite{CC+98}. Finally, another interesting fact is
that graph structures occur almost everywhere in Web mining
research. There are many opportunities for (existing or new)
machine learning algorithms that could work with this
representation or that could take advantage of the available
structures on the Web.

\section{Acknowledgements}

We thank Johannes F\"{u}rnkranz, Dunja Mladenic,  Nico Jacobs, and
Maurice Bruynooghe for reading the draft and the useful comments.
We also thank Luc Dehaspe for the interesting discussions. Raymond
Kosala is supported by the GOA Project LP+ of the Katholieke
Universiteit Leuven. Hendrik Blockeel is a post-doctoral fellow
of the Fund for Scientific Research of Flanders.

\bibliographystyle{abbrv}

\begin{thebibliography}{100}

\bibitem{Abi97}
S.~Abiteboul.
\newblock Querying semi-structured data.
\newblock In {\em Proceedings of Database Theory - ICDT '97, 6th International
  Conference}, volume 1186 of {\em Lecture Notes in Computer Science}, pages
  1--18. Springer, 1997.

\bibitem{AQ+97}
S.~Abiteboul, D.~Quass, J.~McHugh, J.~Widom, and J.~L. Wiener.
\newblock The lorel query language for semistructured data.
\newblock {\em Int. J. on Digital Libraries}, 1(1):68--88, 1997.

\bibitem{AH+98}
H.~Ahonen, O.~Heinonen, M.~Klemettinen, and A.~Verkamo.
\newblock Applying data mining techniques for descriptive phrase extraction in
  digital document collections.
\newblock In {\em Advances in Digital Libraries (ADL'98)}, 1998.

\bibitem{AH+99}
H.~Ahonen, O.~Heinonen, M.~Klemettinen, and A.~Verkamo.
\newblock Finding co-occurring text phrases by combining sequence and frequent
  set discovery.
\newblock In {\em Proceedings of 16th International Joint Conference on
  Artificial Intelligence IJCAI-99 Workshop on Text Mining: Foundations,
  Techniques and Applications}, pages 1--9, 1999.

\bibitem{AC+98}
J.~Allan, J.~Carbonell, G.~Doddington, J.~Yamron, and Y.~Yang.
\newblock Topic detection and tracking pilot study: Final report.
\newblock In {\em Proceedings of the DARPA Broadcast News Transcription and
  Understanding Workshop}, 1998.

\bibitem{APL98}
J.~Allan, R.~Papka, and V.~Lavrenko.
\newblock On-line new event detection and tracking.
\newblock In {\em Proceedings of the 21st annual international ACM SIGIR
  conference on Research and development in information retrieval}, pages
  37--45, 1998.

\bibitem{AI99}
D.~E. Appelt and D.~Israel.
\newblock Introduction to information extraction technology.
\newblock In {\em Proceedings of 16th International Joint Conference on
  Artificial Intelligence IJCAI-99, Tutorial}, 1999.

\bibitem{AM99}
G.~O. Arocena and A.~O. Mendelzon.
\newblock Weboql: Restructuring documents, databases, and webs.
\newblock {\em Theory and Practice of Object Systems}, 5(3):127--141, 1999.

\bibitem{AM97}
P.~Atzeni and G.~Mecca.
\newblock Cut {\&} paste.
\newblock In {\em Proceedings of the Sixteenth ACM SIGACT-SIGMOD-SIGART
  Symposium on Principles of Database Systems}, pages 144--153. ACM Press,
  1997.

\bibitem{YN99}
R.~Baeza-Yates and e.~Berthier Ribeiro-Neto.
\newblock {\em Modern Information Retrieval}.
\newblock Addison-Wesley Longman Publishing Company, 1999.

\bibitem{BS97}
M.~Balabanovic and Y.~Shoham.
\newblock Fab: Content-based, collaborative recommendation.
\newblock {\em Communications of the ACM}, 40(3):66--70, 1997.

\bibitem{BH98}
K.~Bharat and M.~R. Henzinger.
\newblock Improved algorithms for topic distillation in a hyperlinked
  environment.
\newblock In {\em Proceedings of the 21st annual international ACM SIGIR
  conference on Research and development in information retrieval}, pages
  104--111, 1998.

\bibitem{BP99}
D.~Billsus and M.~Pazzani.
\newblock A hybrid user model for news story classification.
\newblock In {\em Proceedings of the Seventh International Conference on User
  Modeling (UM '99)}, 1999.

\bibitem{BL98}
J.~Borges and M.~Levene.
\newblock Mining association rules in hypertext databases.
\newblock In {\em Proceedings of the Fourth International Conference on
  Knowledge Discovery and Data Mining (KDD-98)}, 1998.

\bibitem{BL99}
J.~Borges and M.~Levene.
\newblock Data mining of user navigation patterns.
\newblock In {\em Proceedings of the WEBKDD'99 Workshop on Web Usage Analysis
  and User Profiling}, pages 31--36, 1999.

\bibitem{BP98}
S.~Brin and L.~Page.
\newblock The anatomy of a large-scale hypertextual {Web} search engine.
\newblock In {\em Seventh International World Wide Web Conference}, 1998.

\bibitem{BB+99}
A.~B\"{u}chner, M.~Baumgarten, S.~Anand, M.~Mulvenna, and J.~Hughes.
\newblock Navigation pattern discovery from internet data.
\newblock In {\em Proceedings of the WEBKDD'99 Workshop on Web Usage Analysis
  and User Profiling}, 1999.

\bibitem{Bun97}
P.~Buneman.
\newblock Semistructured data.
\newblock In {\em Proceedings of the Sixteenth ACM SIGACT-SIGMOD-SIGART
  Symposium on Principles of Database Systems}, pages 117--121. ACM Press,
  1997.

\bibitem{BD+96}
P.~Buneman, S.~B. Davidson, G.~G. Hillebrand, and D.~Suciu.
\newblock A query language and optimization techniques for unstructured data.
\newblock In {\em Proceedings of the 1996 ACM SIGMOD International Conference
  on Management of Data}, pages 505--516. ACM Press, 1996.

\bibitem{CC+98}
J.~Carbonell, M.~Craven, S.~Fienberg, T.~Mitchell, and Y.~Yang.
\newblock Report on the conald workshop on learning from text and the web.
\newblock In {\em CONALD Workshop on Learning from Text and the Web}, 1998.

\bibitem{CYC00}
J.~Carbonell, Y.~Yang, and W.~Cohen.
\newblock Special issue of machine learning on information retrieval
  introduction.
\newblock {\em Machine Learning}, 39:99--101, 2000.

\bibitem{Car97}
C.~Cardie.
\newblock Empirical methods in information extraction.
\newblock {\em AI Magazine}, 18(4):65--79, 1997.

\bibitem{Cha00}
S.~Chakrabarti.
\newblock Data mining for hypertext: A tutorial survey.
\newblock {\em ACM SIGKDD Explorations}, 1(2):1--11, 2000.

\bibitem{CD+99}
S.~Chakrabarti, B.~Dom, D.~Gibson, J.~Kleinberg, S.~Kumar,
P.~Raghavan,
  S.~Rajagopalan, and A.~Tomkins.
\newblock Mining the link structure of the world wide web.
\newblock {\em IEEE Computer}, 32(8):60--67, 1999.

\bibitem{CDI98}
S.~Chakrabarti, B.~Dom, and P.~Indyk.
\newblock Enhanced hypertext categorization using hyperlinks.
\newblock In {\em SIGMOD 1998, Proceedings ACM SIGMOD International Conference
  on Management of Data}, pages 307--318. ACM Press, 1998.

\bibitem{CM+94}
S.~Chawathe, H.~Garcia-Molina, J.~Hammer, K.~Ireland,
Y.~Papakonstantinou,
  J.~Ullman, and J.~Widom.
\newblock The tsimmis project: Integration of heterogeneous information
  sources.
\newblock In {\em Proceedings of the 10th Meeting of the Information Processing
  Society of Japan}, pages 7--18, 1994.

\bibitem{Coh95}
W.~W. Cohen.
\newblock Learning to classify english text with ilp methods.
\newblock In {\em Advances in Inductive Logic Programming (Ed. L. De Raedt)}.
  IOS Press, 1995.

\bibitem{Coh99b}
W.~W. Cohen.
\newblock Some practical observations on integration of web information.
\newblock In {\em ACM SIGMOD Workshop on The Web and Databases (WebDB'99)},
  pages 55--60, Philadelphia, Pennsylvania, USA, 1999.

\bibitem{Coh99a}
W.~W. Cohen.
\newblock What can we learn from the web?
\newblock In {\em Proceedings of the Sixteenth International Conference on
  Machine Learning (ICML'99)}, pages 515--521, 1999.

\bibitem{CMS97}
R.~Cooley, B.~Mobasher, and J.~Srivastava.
\newblock Web mining: Information and pattern discovery on the world wide web.
\newblock In {\em Proceedings of the 9th IEEE International Conference on Tools
  with Artificial Intelligence (ICTAI'97)}, 1997.

\bibitem{CMS99}
R.~Cooley, B.~Mobasher, and J.~Srivastava.
\newblock Data preparation for mining world wide web browsing patterns.
\newblock {\em Knowledge and Information Systems}, 1(1), 1999.

\bibitem{Coo00}
R.~W. Cooley.
\newblock {\em Web Usage Mining: Discovery and Application of Interesting
  Patterns from Web data}.
\newblock PhD thesis, Dept. of Computer Science, University of Minnesota, May
  2000.

\bibitem{CL96}
J.~Cowie and W.~Lehnert.
\newblock Information extraction.
\newblock {\em Communications of the ACM}, 39(1):80--91, 1996.

\bibitem{CD+98}
M.~Craven, D.~DiPasquo, D.~Freitag, A.~McCallum, T.~Mitchell,
K.~Nigam, and
  S.~Slattery.
\newblock Learning to extract symbolic knowledge from the world wide web.
\newblock In {\em Proceedings of the Fifteenth National Conference on
  Artificial Intellligence (AAAI98)}, pages 509--516, 1998.

\bibitem{CS+99}
F.~Crimmins, A.~Smeaton, T.~Dkaki, and J.~Mothe.
\newblock T\'{e}trafusion: Information discovery on the internet.
\newblock {\em IEEE Intelligent Systems}, 14(4):55--62, 1999.

\bibitem{DD+90}
S.~Deerwester, S.~Dumais, G.~Furnas, T.~Landauer, and R.~Harshman.
\newblock Indexing by latent semantic analysis.
\newblock {\em Journal of the American Society for Information Science},
  41(6):391--407, 1990.

\bibitem{DD97}
L.~Dehaspe and L.~de~Raedt.
\newblock Mining association rules in multiple relations.
\newblock In {\em Proceedings of the 7th International Workshop on Inductive
  Logic Programming}, volume 1297 of {\em Lecture Notes in Computer Science},
  pages 125--132, Prague, Czech Republic, 1997. Springer.

\bibitem{Del00}
J.~A. Delgado.
\newblock {\em Agent-Based Information Filtering and Recommender System On the
  Internet}.
\newblock PhD thesis, Dept. of Intelligence \& Computer Science, Nagoya
  Institute of Technology, March 2000.

\bibitem{DP+98}
S.~Dumais, J.~Platt, D.~Heckerman, and M.~Sahami.
\newblock Inductive learning algorithms and representations for text
  categorization.
\newblock In {\em Proceedings of the 1998 ACM 7th international conference on
  Information and knowledge management}, pages 148--155, Washington United
  States, 1998.

\bibitem{Etz96}
O.~Etzioni.
\newblock The world wide web: Quagmire or gold mine.
\newblock {\em Communications of the ACM}, 39(11):65--68, 1996.

\bibitem{FDW96}
U.~Fayyad, S.~Djorgovski, and N.~Weir.
\newblock Automating the analysis and cataloging of sky surveys.
\newblock In {\em Advances in Knowledge Discovery and Data Mining}, pages
  471--493. AAAI Press, 1996.

\bibitem{FSS96b}
U.~Fayyad, G.~Piatetsky-Shapiro, and P.~Smyth.
\newblock From data mining to knowledge discovery: An overview.
\newblock In {\em Advances in Knowledge Discovery and Data Mining}, pages
  1--34. AAAI Press, 1996.

\bibitem{FSS96a}
U.~Fayyad, G.~Piatetsky-Shapiro, and P.~Smyth.
\newblock Knowledge discovery and data mining: toward a unifying framework.
\newblock In {\em Proceeding of The Second Int. Conference on Knowledge
  Discovery and Data Mining}, pages 82--88, 1996.

\bibitem{FD95}
R.~Feldman and I.~Dagan.
\newblock Knowledge discovery in textual databases (kdt).
\newblock In {\em Proceedings of the First International Conference on
  Knowledge Discovery and Data Mining (KDD-95)}, pages 112--117, Montreal,
  Canada, 1995.

\bibitem{FF+98}
R.~Feldman, M.~Fresko, Y.~Kinar, Y.~Lindell, O.~Liphstat, M.~Rajman,
Y.~Schler,
  and O.~Zamir.
\newblock Text mining at the term level.
\newblock In {\em Principles of Data Mining and Knowledge Discovery, Second
  European Symposium, PKDD '98}, volume 1510 of {\em Lecture Notes in Computer
  Science}, pages 56--64. Springer, 1998.

\bibitem{FK+00}
D.~Fensel, C.~Knoblock, N.~Kushmerick, and M.-C. Rousset.
\newblock Workshop on intelligent information integration (iii'99).
\newblock {\em AI Magazine}, 21(1):91--94, 2000.

\bibitem{FF+97}
M.~F. Fernandez, D.~Florescu, A.~Y. Levy, and D.~Suciu.
\newblock A query language for a web-site management system.
\newblock {\em SIGMOD Record}, 26(3):4--11, 1997.

\bibitem{FP99}
R.~E. Filman and S.~Pant.
\newblock Searching the internet - guest editors' introduction.
\newblock {\em IEEE Internet Computing}, 2(4):21--23, 1998.

\bibitem{FLM98}
D.~Florescu, A.~Y. Levy, and A.~O. Mendelzon.
\newblock Database techniques for the world-wide web: A survey.
\newblock {\em SIGMOD Record}, 27(3):59--74, 1998.

\bibitem{FP+99}
E.~Frank, G.~W. Paynter, I.~H. Witten, C.~Gutwin, and C.~G.
Nevill-Manning.
\newblock Domain-specific keyphrase extraction.
\newblock In {\em Proceedings of 16th International Joint Conference on
  Artificial Intelligence IJCAI-99}, pages 668--673, 1999.

\bibitem{Fre98}
D.~Freitag.
\newblock Information extraction from html: Application of a general learning
  approach.
\newblock In {\em Proceedings of the Fifteenth Conference on Artificial
  Intelligence AAAI-98}, pages 517--523, 1998.

\bibitem{FM99}
D.~Freitag and A.~McCallum.
\newblock Information extraction with hmms and shrinkage.
\newblock In {\em Proceedings of the AAAI-99 Workshop on Machine Learning for
  Information Extraction}, 1999.

\bibitem{Fur99}
J.~F\"{u}rnkranz.
\newblock Exploiting structural information for text classification on the www.
\newblock In {\em Advances in Intelligent Data Analysis, Third International
  Symposium, IDA-99}, pages 487--498, 1999.

\bibitem{GR+99}
M.~N. Garofalakis, R.~Rastogi, S.~Seshadri, and K.~Shim.
\newblock Data mining and the web: Past, present and future.
\newblock In {\em Workshop on Web Information and Data Management, 1999}, pages
  43--47, 1999.

\bibitem{GW97}
R.~Goldman and J.~Widom.
\newblock Dataguides: Enabling query formulation and optimization in
  semistructured databases.
\newblock In {\em VLDB'97, Proceedings of 23rd International Conference on Very
  Large Data Bases}, pages 436--445. Morgan Kaufmann, 1997.

\bibitem{GW99}
R.~Goldman and J.~Widom.
\newblock Approximate dataguides.
\newblock In {\em Proceedings of the Workshop on Query Processing for
  Semistructured Data and Non-Standard Data Formats}, 1999.

\bibitem{GH+97}
S.~Green, L.~Hurst, B.~Nangle, P.~Cunningham, F.~Somers, and
R.~Evans.
\newblock Software agents: A review.
\newblock Technical Report TCD-CS-1997-06, Technical Report of Trinity College,
  University of Dublin, 1997.

\bibitem{GM99}
S.~Grumbach and G.~Mecca.
\newblock In search of the lost schema.
\newblock In {\em Database Theory - ICDT '99, 7th International Conference},
  pages 314--331, 1999.

\bibitem{HG+97}
J.~Hammer, H.~Garcia-Molina, J.~Cho, A.~Crespo, and R.~Aranha.
\newblock Extracting semistructured information from the web.
\newblock In {\em Proceedings of the Workshop on Management of Semistructured
  Data}, pages 18--25, 1997.

\bibitem{Hau99}
A.~Hauptmann.
\newblock Integrating and using large databases of text, image, video and
  audio.
\newblock {\em IEEE Intelligent Systems}, 14(5):34--35, 1999.

\bibitem{Hea99}
M.~A. Hearst.
\newblock Untangling text data mining.
\newblock In {\em Proceedings of ACL'99: the 37th Annual Meeting of the
  Association for Computational Linguistics}, 1999.

\bibitem{Hof99}
T.~Hofmann.
\newblock The cluster-abstraction model: Unsupervised learning of topic
  hierarchies from text data.
\newblock In {\em Proceedings of 16th International Joint Conference on
  Artificial Intelligence IJCAI-99}, pages 682--687, 1999.

\bibitem{HW99}
S.~J. Hong and S.~M. Weiss.
\newblock Advances in predictive model generation for data mining.
\newblock Technical Report Report RC-21570, IBM Research Report, 1999.

\bibitem{HK97}
T.~Honkela, S.~Kaski, K.~Lagus, and T.~Kohonen.
\newblock Websom - self-organizing maps of document collections.
\newblock In {\em Proc. of Workshop on Self-Organizing Maps (WSOM'97)}, pages
  310--315, 1997.

\bibitem{HC+99}
A.~Houston, H.~Chen, S.~M. Hubbard, B.~R. Schatz, T.~D. Ng, R.~R.
Sewell, and
  K.~M. Tolle.
\newblock Medical data mining on the internet: Research on a cancer information
  system.
\newblock {\em Artificial Intelligence Review}, 13:437--446, 1999.

\bibitem{HD98}
C.-N. Hsu and M.-T. Dung.
\newblock Generating finite-state transducers for semi-structured data
  extraction from the web.
\newblock {\em Information Systems}, 23(8):521--538, 1998.

\bibitem{JFM97}
T.~Joachims, D.~Freitag, and T.~Mitchell.
\newblock Webwatcher: A tour guide for the world wide web.
\newblock In {\em Proceedings of the International Joint Conference on
  Artificial Intelligence IJCAI-97}, pages 770--777, 1997.

\bibitem{JSR99}
M.~Junker, M.~Sintek, and M.~Rinck.
\newblock Learning for text categorization and information extraction with ilp.
\newblock In {\em Proceedings of the Workshop on Learning Language in Logic},
  1999.

\bibitem{WL99}
H.~L. K.~Wang.
\newblock Discovering association of structure from semistructured objects.
\newblock {\em To appear in IEEE Transactions on Knowledge and Data
  Engineering}, 1999.

\bibitem{KHS97}
H.~Kargupta, I.~Hamzaoglu, and B.~Stafford.
\newblock Distributed data mining using an agent based architecture.
\newblock In {\em Proceedings of Knowledge Discovery And Data Mining}, pages
  211--214. AAAI Press, 1997.

\bibitem{KSS97}
H.~Kautz, B.~Selman, and M.~Shah.
\newblock The hidden web.
\newblock {\em AI magazine}, 18(2):27--36, 1997.

\bibitem{KB96}
S.~Khoshafian and A.~B. Baker.
\newblock {\em Multimedia and Imaging Databases}.
\newblock Morgan Kaufmann Publishers, 1996.

\bibitem{Kle98}
J.~M. Kleinberg.
\newblock Authoritative sources in a hyperlinked environment.
\newblock In {\em Proc. of ACM-SIAM Symposium on Discrete Algorithms}, pages
  668--677, 1998.

\bibitem{Kod99}
Y.~Kodratoff.
\newblock About knowledge discovery in texts: A definition and an example.
\newblock In {\em Proc. of Advanced Course on Artificial Intelligence (ACAI-99)
  on Machine Learning and Applications (Invited talk)}, 1999.

\bibitem{KR+98}
S.~R. Kumar, P.~Raghavan, S.~Rajagopalan, and A.~Tomkins.
\newblock Trawling the web for emerging cyber-communities.
\newblock In {\em Proceedings of the Eighth World Wide Web Conference (WWW8)},
  1999.

\bibitem{Kus99}
N.~Kushmerick.
\newblock Gleaning the web.
\newblock {\em IEEE Intelligent Systems}, 14(2):20--22, 1999.

\bibitem{KWD97}
N.~Kushmerick, D.~Weld, and R.~Doorenbos.
\newblock Wrapper induction for information extraction.
\newblock In {\em Proceedings of the International Joint Conference on
  Artificial Intelligence IJCAI-97}, pages 729--737, 1997.

\bibitem{LSS96}
L.~Lakshmanan, F.~Sadri, and I.~Subramanian.
\newblock A declarative language for querying and restructuring the web.
\newblock In {\em Proceedings of 6th. International Workshop on Research Issues
  in Data Engineering, RIDE '96}, pages 12--21, 1996.

\bibitem{Lan99}
P.~Langley.
\newblock User modeling in adaptive interfaces.
\newblock In {\em Proceedings of the Seventh International Conference on User
  Modeling}, pages 357--370, 1999.

\bibitem{LG99}
S.~Lawrence and C.~L. Giles.
\newblock Accessibility of information on the web.
\newblock {\em Nature}, 400:107--109, 1999.

\bibitem{LAS97}
B.~Lent, R.~Agrawal, and R.~Srikant.
\newblock Discovering trends in text databases.
\newblock In {\em Proc. 3 rd Int Conf. On Knowledge Discovery and Data Mining
  (KDD 1997)}, pages 227--230, 1997.

\bibitem{LW00}
A.~Y. Levy and D.~S. Weld.
\newblock Intelligent internet systems.
\newblock {\em Artificial Intelligence}, 118(1-2), 2000.

\bibitem{MB+99}
S.~K. Madria, S.~S. Bhowmick, W.~K. Ng, and E.-P. Lim.
\newblock Research issues in web data mining.
\newblock In {\em Proceedings of Data Warehousing and Knowledge Discovery,
  First International Conference, DaWaK '99}, pages 303--312, 1999.

\bibitem{Mae94}
P.~Maes.
\newblock Agents that reduce work and information overload.
\newblock {\em Communications of the ACM}, 37(7):30--40, 1994.

\bibitem{MS00}
B.~Masand and M.~Spiliopoulou.
\newblock Webkdd-99: Workshop on web usage analysis and user profiling.
\newblock {\em ACM SIGKDD Explorations}, 1(2), 2000.

\bibitem{MN+99}
A.~McCallum, K.~Nigam, J.~Rennie, and K.~Seymore.
\newblock A machine learning approach to building domain-specific search
  engines.
\newblock In {\em Proceedings of the International Joint Conference on
  Artificial Intelligence IJCAI-99}, pages 662--667, 1999.

\bibitem{MMM96}
A.~O. Mendelzon, G.~A. Mihaila, and T.~Milo.
\newblock Querying the world wide web.
\newblock In {\em Proceedings of the Fourth International Conference on
  Parallel and Distributed Information Systems}, pages 80--91, 1996.

\bibitem{Mit97}
T.~Mitchell.
\newblock {\em Machine Learning}.
\newblock McGraw Hill, 1997.

\bibitem{Mit99}
T.~M. Mitchell.
\newblock Machine learning and data mining.
\newblock {\em Communications of the ACM}, 42(11):30--36, 1999.

\bibitem{Mla99}
D.~Mladenic.
\newblock Text-learning and related intelligent agents.
\newblock {\em IEEE Intelligent Systems}, 14(4):44--54, 1999.

\bibitem{MG99}
D.~Mladenic and M.~Grobelnik.
\newblock Feature selection for unbalanced class distribution and naive bayes.
\newblock In {\em Proceedings of the 16th International Conference on Machine
  Learning ICML-99}, pages 258--267, 1999.

\bibitem{Mus99}
I.~Muslea.
\newblock Extraction patterns for information extraction tasks: A survey.
\newblock In {\em AAAI-99 Workshop on Machine Learning for Information
  Extraction}, 1999.

\bibitem{MMK98}
I.~Muslea, S.~Minton, and C.~Knoblock.
\newblock Wrapper induction for semistructured, web-based information sources.
\newblock In {\em Proceedings of the Conference on Automatic Learning and
  Discovery CONALD-98}, 1998.

\bibitem{NM00}
U.~Y. Nahm and R.~J. Mooney.
\newblock A mutually beneficial integration of data mining and information
  extraction.
\newblock In {\em Proceedings of the Seventeenth National Conference on
  Artificial Intelligence (AAAI-00)}, 2000.

\bibitem{NAM97}
S.~Nestorov, S.~Abiteboul, and R.~Motwani.
\newblock Infering structure in semistructured data.
\newblock {\em SIGMOD Record}, 26(4), 1997.

\bibitem{NAM98}
S.~Nestorov, S.~Abiteboul, and R.~Motwani.
\newblock Extracting schema from semistructured data.
\newblock In L.~M. Haas and A.~Tiwary, editors, {\em SIGMOD 1998, Proceedings
  ACM SIGMOD International Conference on Management of Data}, pages 295--306.
  ACM Press, 1998.

\bibitem{NLM99}
K.~Nigam, J.~Lafferty, and A.~McCallum.
\newblock Using maximum entropy for text classification.
\newblock In {\em Proceedings of the International Joint Conference on
  Artificial Intelligence IJCAI-99 Workshop on Machine Learning for Information
  Filtering}, pages 61--67, 1999.

\bibitem{PP+00}
G.~Paliouras, C.~Papatheodorou, V.~Karkaletsis, P.~Tzitziras, and
C.~D.
  Spyropoulos.
\newblock Large-scale mining of usage data on web sites.
\newblock In {\em AAAI 2000 Spring Symposium on Adaptive User Interfaces},
  2000.

\bibitem{Paz97}
M.~T. Pazienza, editor.
\newblock {\em Information Extraction: A multidisciplinary Approach to an
  Emerging Information Technology}, volume 1299 of {\em Lecture Notes in
  Computer Science}. International Summer School, SCIE-97, Frascati (Rome),
  Springer, 1997.

\bibitem{Paz99}
M.~T. Pazienza, editor.
\newblock {\em Information Extraction}, Frascati (Rome), 1999. International
  Summer School, SCIE-99 , Frascati (Rome).

\bibitem{SB+96}
G.~Piatetsky-Shapiro, R.~Braachman, T.~Khabaza, W.~Kloesgen, and
E.~Simoudis.
\newblock An overview of issues in developing industrial data mining and
  knowledge discovery applications.
\newblock In {\em Proceeding of the Second Int. Conference on Knowledge
  Discovery and Data Mining (KDD-96)}, pages 89--95, 1996.

\bibitem{RB98}
M.~Rajman and R.~Besan\c{c}on.
\newblock Text mining - knowledge extraction from unstructured textual data.
\newblock In {\em Proc. of 6th Conference of International Federation of
  Classification Societies (IFCS-98)}, pages 473--480, 1998.

\bibitem{RM99b}
A.~Rauber and D.~Merkl.
\newblock Automatic labeling of self-organizing maps: Making a treasure-map
  reveal its secrets.
\newblock In {\em Proc of the Pacific Asia Conf on Knowledge Discovery and Data
  Mining (PAKDD'99)}, 1999.

\bibitem{RM99a}
J.~Rennie and A.~McCallum.
\newblock Using reinforcement learning to spider the web efficiently.
\newblock In {\em Proceedings of the 16th International Conference on Machine
  Learning ICML-99}, 1999.

\bibitem{Ril95}
E.~Riloff.
\newblock Little words can make a big difference for text classification.
\newblock In {\em SIGIR'95, Proceedings of the 18th Annual International ACM
  SIGIR Conference on Research and Development in Information Retrieval}, pages
  130--136. ACM Press, 1995.

\bibitem{SM83}
G.~Salton and M.~McGill.
\newblock {\em Introduction to Modern Information Retrieval}.
\newblock McGraw Hill, 1983.

\bibitem{SM99}
S.~Scott and S.~Matwin.
\newblock Feature engineering for text classification.
\newblock In {\em Proceedings of the 16th International Conference on Machine
  Learning ICML-99}, 1999.

\bibitem{SR98}
J.~W. Shavlik and T.~Eliassi-Rad.
\newblock Intelligent agents for web-based tasks: An advice-taking approach.
\newblock In {\em Working Notes of the AAAI/ICML-98 Workshop on Learning for
  Text Categorization}, pages 588--589, 1999.

\bibitem{SC+98}
L.~Singh, B.~Chen, R.~Haight, P.~Scheuermann, and K.~Aoki.
\newblock A robust system architecture for mining semi-structured data.
\newblock In {\em Proceeding of The Fourth Int. Conference on Knowledge
  Discovery and Data Mining (KDD-98)}, pages 329--333, 1998.

\bibitem{SF+96}
P.~Smyth, U.~M. Fayyad, M.~C. Burl, and P.~Perona.
\newblock Modeling subjective uncertainty in image annotation.
\newblock {\em Advances in Knowledge Discovery and Data Mining}, pages
  517--539, 1996.

\bibitem{Sod99}
S.~Soderland.
\newblock Learning information extraction rules for semi-structured and free
  text.
\newblock {\em Machine Learning}, 34(1-3):233--272, 1999.

\bibitem{Spi99}
M.~Spiliopoulou.
\newblock Data mining for the web.
\newblock In {\em Principles of Data Mining and Knowledge Discovery, Second
  European Symposium, PKDD '99}, pages 588--589, 1999.

\bibitem{SC+00}
J.~Srivastava, R.~Cooley, M.~Deshpande, and P.-N. Tan.
\newblock Web usage mining: Discovery and applications of usage patterns from
  web data.
\newblock {\em SIGKDD Explorations}, 1(2), 2000.

\bibitem{Sub98}
V.~S. Subrahmanian.
\newblock {\em Principles of Multimedia Database Systems}.
\newblock Morgan Kaufmann Publishers, 1998.

\bibitem{Tan99}
A.-H. Tan.
\newblock Text mining: The state of the art and the challenges.
\newblock In {\em Proc of the Pacific Asia Conf on Knowledge Discovery and Data
  Mining PAKDD'99 workshop on Knowledge Discovery from Advanced Databases},
  pages 65--70, 1999.

\bibitem{Toi99}
H.~Toivonen.
\newblock On knowledge discovery in graph-structured data.
\newblock In {\em Workshop on Knowledge Discovery from Advanced Databases
  (KDAD'99)}, pages 26--31, 1999.

\bibitem{Uth96}
R.~Uthurusamy.
\newblock From data mining to knowledge discovery: Current challenges and
  future directions.
\newblock In {\em Advances in Knowledge Discovery and Data Mining}, pages
  561--569, 1996.

\bibitem{Vai99}
S.~Vaithyanathan.
\newblock Introduction: Data mining on the internet.
\newblock {\em Artificial Intelligence Review}, 13(5/6):343--344, 1999.

\bibitem{Van79}
C.~J. van Rijsbergen.
\newblock {\em Information Retrieval}.
\newblock Butterworths, 1979.

\bibitem{WL97}
K.~Wang and H.~Liu.
\newblock Schema discovery for semistructured data.
\newblock In {\em Proceedings of the Third International Conference on
  Knowledge Discovery and Data Mining (KDD-97)}, pages 271--274, 1997.

\bibitem{WA+99}
S.~M. Weiss, C.~Apt\'{e}, F.~Damerau, D.~E. Johnson, F.~J. Oles,
T.~Goetz, and
  T.~Hampp.
\newblock Maximizing text-mining performance.
\newblock {\em IEEE Intelligent Systems}, 14(4):63--69, 1999.

\bibitem{WPW95}
W.~Wiener, J.~Pedersen, and A.~Weigend.
\newblock A neural network approach to topic spotting.
\newblock In {\em Proceedings of the 4th Symposium on Document Analysis and
  Information Retrieval (SDAIR 95)}, pages 317--332, 1995.

\bibitem{Wil97}
Y.~Wilks.
\newblock {\em Information Extraction as a core language technology}, volume
  1299 of {\em Lecture Notes in Computer Science}, chapter In M-T. Pazienza
  (ed.), Information Extraction, pages 1--9.
\newblock Springer, 1997.

\bibitem{WB+99}
I.~H. Witten, Z.~Bray, M.~Mahoui, and W.~J. Teahan.
\newblock Text mining: A new frontier for lossless compression.
\newblock In {\em Data Compression Conference}, pages 198--207, 1999.

\bibitem{YC+99}
Y.~Yang, J.~Carbonell, R.~Brown, T.~Pierce, B.~T. Archibald, and
X.~Liu.
\newblock Learning approaches for detecting and tracking news events.
\newblock {\em IEEE Intelligent Systems}, 14(4):32--43, 1999.

\bibitem{YP99}
Y.~Yang and J.~Pedersen.
\newblock Guest editors' introduction: Intelligent information retrieval.
\newblock {\em IEEE Intelligent Systems}, 14(4):30--31, 1999.

\bibitem{ZH98}
O.~Zaiane and J.~Han.
\newblock Webml: Querying the world-wide web for resources and knowledge.
\newblock In {\em Proc. ACM CIKM'98 Workshop on Web Information and Data
  Management (WIDM'98)}, pages 9--12, 1998.

\bibitem{ZH+98}
O.~R. Zaiane, J.~Han, Z.-N. Li, S.~H. Chee, and J.~Chiang.
\newblock Multimediaminer: a system prototype for multimedia data mining.
\newblock In {\em Proc. ACM SIGMOD Intl. Conf. on Management of Data}, pages
  581--583, 1998.

\end{thebibliography}

\end{document}